\documentclass[conference]{IEEEtran}
\IEEEoverridecommandlockouts
\usepackage{amsmath,amssymb,amsfonts}
\usepackage{algorithmic}
\usepackage{graphicx}
\usepackage{textcomp}
\usepackage{xcolor}

\usepackage{booktabs}
\usepackage{tabu}
\usepackage{hyperref} 
\usepackage{multirow}
\usepackage[numbers]{natbib}

\def\BibTeX{{\rm B\kern-.05em{\sc i\kern-.025em b}\kern-.08em
    T\kern-.1667em\lower.7ex\hbox{E}\kern-.125emX}}

\makeatletter
\def\ps@IEEEtitlepagestyle{%
  \def\@oddfoot{\footnotesize\hfil
    \begin{tabular}{c}
      IECON 2024 - 50th Annual Conference of the IEEE Industrial Electronics Society \\[-0.4ex]
      IEEE Copyright \url{https://www.ieee.org/publications/rights/copyright-policy.html}
    \end{tabular}%
    \hfil}%
  \def\@evenfoot{\@oddfoot}%
}
\makeatother
    
\begin{document}

\title{Graph Attention Convolutional U-NET: A Semantic Segmentation Model for Identifying Flooded Areas \thanks{{This research has been funded by NSERC under grant RGPIN-2018-06222 (K. Grolinger) and by the Vector Scholarship in Artificial Intelligence, Vector Institute (M. Buwaneswaran). Computation was enabled in part by the Digital Research Alliance of Canada.}}}

\author{
    \IEEEauthorblockN{Muhammad Umair Danish, Madhushan Buwaneswaran, Tehara Fonseka, Katarina Grolinger}
    \IEEEauthorblockA{Department of Electrical and Computer Engineering, Western University, London, Ontario, Canada}
    \{mdanish3, mbuwanes, tfonsek, kgroling\}@uwo.ca
}

\maketitle

\begin{abstract}

The increasing impact of human-induced climate change and unplanned urban constructions has increased flooding incidents in recent years. Accurate identification of flooded areas is crucial for effective disaster management and urban planning. While few works have utilized convolutional neural networks and transformer-based semantic segmentation techniques for identifying flooded areas from aerial footage, recent developments in graph neural networks have created improvement opportunities. This paper proposes an innovative approach, the Graph Attention Convolutional U-NET (GAC-UNET) model, based on graph neural networks for automated identification of flooded areas. The model incorporates a graph attention mechanism and Chebyshev layers into the U-Net architecture. Furthermore, this paper explores the applicability of transfer learning and model reprogramming to enhance the accuracy of flood area segmentation models. Empirical results demonstrate that the proposed GAC-UNET model, outperforms other approaches with 91\% mAP, 94\% dice score, and 89\% IoU, providing valuable insights for informed decision-making and better planning of future infrastructures in flood-prone areas. 

\end{abstract}


\section{Introduction} \label{sec:intro}

Human-induced climate change \cite{alifu2022enhancement} and unplanned urban constructions \cite{andreasen2023built} have increased flooding more prominently in recent years. Flood surveys are required to identify these flooded areas correctly. During floods, aerial footage is obtained through aerial vehicles. This footage can be analyzed to determine the extent of the flood. Accurately quantifying the extent of flooding and identifying areas prone to repeated flooding can provide valuable insights for informed decision-making and better planning of future infrastructures in those areas. This approach can help mitigate potential risks and ensure the community's safety and well-being.

A possible way of automating the identification of flooded areas from images is through semantic segmentation which is a task in computer vision that involves partitioning an image into multiple regions. Specifically, semantic segmentation generates a dense pixel-wise segmentation map of an image by assigning each pixel in the image to a class, thus providing a precise segmentation of objects present in an image. The applications of semantic segmentation are widespread, especially in areas such as medical image processing and autonomous driving. For example, in the medical domain, semantic segmentation is used for several use cases such as identifying pathology location, quantifying tissue volumes, and studying anatomical structure. Traditionally semantic segmentation was done using methods such as thresholding, edge detection, and clustering. However, recently machine learning approaches, specifically deep learning techniques have become state-of-the-art methods for semantic segmentation.

Flood area segmentation is modelled as a two-class (binary) semantic segmentation problem where flood area pixels are labelled as ones and the remaining regions are labelled as zeros. Few works have studied the problem of flood detection through semantic segmentation \cite{muhadi2021deep, sanderson2023optimal, roy2022transformer}. These works utilized deep learning techniques such as convolutional neural networks and transformers. While these works have improved the semantic segmentation of flooded regions, graph neural networks, a recent advancement in deep learning, provide an opportunity to further improve flood detection. Utilizing nodes and edges to capture relationships, graph networks enable learning from data with complex relational structures. In recent years, they have shown promising results in related domains such as medical image segmentation \cite{van2022finite} and scene segmentation in video \cite{varga2021fast}.

Furthermore, Machine Learning (ML) techniques such as transfer learning and model reprogramming have demonstrated the capability to improve the performance of deep learning models on downstream tasks with limited data availability by distilling knowledge from one task to another. While transfer learning transfers knowledge from a pre-trained model to another task and fine-tunes the model on the second dataset, reprogramming enables re-purposing a pre-trained model without fine-tuning, by introducing input transformation and output mapping \cite{chen2022model}. These techniques have been successfully applied in computer vision, natural language processing, and time series domains \cite{chen2022model}. Transfer learning and model reprogramming also offer opportunities to improve the performance of flood area segmentation models.

Consequently, this paper proposes a novel flood area identification approach, Graph Attention Convolutional U-NET (GAC-UNET), based on graph attention networks—a graph neural network variant that incorporates an attention mechanism—-due to their capability to learn complex spatial relationships \footnote{ Code can be found at: https://github.com/bmsknight/flood-segmentation/}. Chebyshev graph convolutional layer and the center of mass layer are added to improve performance. Furthermore, this paper studies the applicability of transfer learning and model reprogramming to the problem of flood area segmentation. We empirically show that our proposed model GAC-UNET coupled with the dice loss outperforms other considered approaches. Moreover, we demonstrate that both transfer learning and model reprogramming increase accuracy, with transfer learning yielding a greater improvement.

The remainder of the paper is organized as follows: Section \ref{sec:related} explains related works, Section \ref{sec:sota} gives background and preliminaries, Section \ref{sec:method} explains the proposed methodology, and Section \ref{sec:results} presents the overall results and analysis. Finally, Section \ref{sec:conclusion} concludes the paper.

\section{Related Work} \label{sec:related}

This section first reviews recent studies in semantic segmentation in flood monitoring. Next, core transfer learning and model reprogramming works are discussed. Finally, as we employ graph networks in our approach, the prominent works in graph network works are presented.

The highly accurate and efficient flood monitoring systems have increasingly been utilizing deep learning (DL) methods \cite{xie2021segformer}, such as convolutional neural network (CNN)-based architectures were very popular for semantic segmentation before being overtaken by U-Net-based architectures \cite{U-Net}. The introduction of skip connections between the encoder and decoder further substantially improved the performance of U-Net for semantic segmentation \cite{maskrcnn}. Rafi et al. \cite{sanderson2023optimal} proposed an explainable deep CNN that utilizes multi-spectral optical and Synthetic Aperture Radar (SAR) images for flood inundation mapping.  Mahadi et al. proposed a U-Net-like hybrid model namely DeepLabv3+ that also applied atrous convolution for water region segmentation from surveillance footage \cite{muhadi2021deep} that outperformed U-Net. Hern{\'a}ndez at al. \cite{relw2} utilized a U-Net-like model with unmanned aerial vehicles (UAVs) equipped with on-board edge computing to process flood-related data locally, consequently enabling faster response times and reducing dependency on distant computational resources. Recently transformer-based architectures have been used for segmentation as well \cite{xie2021segformer}. Another similar study by Roy et al. \cite{roy2022transformer} introduced FloodTransformer, a transformer-based model specifically designed for segmenting flood scenes from aerial images, which demonstrated outstanding performance across several benchmarks.  

Transfer learning has been used in multiple applications to improve performance when limited data is available. Wu et al. \cite{wu2023near} utilized transfer learning to adapt pre-trained deep learning models to the task of near-real-time flood detection using Synthetic Aperture Radar (SAR) images. Another notable work by Ghosh et al. \cite{ghosh2022automatic} applied transfer learning to fine-tune CNN-based architecture which is optimized for analyzing image data, resulting in substantial enhancements in automatic flood detection capabilities. Recently, a new concept of model reprogramming emerged. Reprogramming enables re-purposing a pre-trained model without fine-tuning by introducing input transformation and output mapping\cite{chen2022model}. Model reprogramming has been successfully applied in speech, computer vision, NLP, and time series domain \cite{chen2022model}.

Graph Neural Networks (GNNs) have shown substantial promise in various domains. For instance, in the field of medical imaging, ViG-UNet integrates vision graph neural networks to enhance medical image segmentation by showcasing potential pathways for similar adaptations in environmental scenes \cite{van2022finite}. Another study explored unsupervised image segmentation using GNNs, which maximizes mutual information to achieve segmentation without labelled data, a method that could dramatically reduce the need for annotated flood images \cite{wong2021meta}. Furthermore, a novel approach employing GNNs for dynamic scene segmentation in videos presented methodologies that could be adapted for analyzing temporal changes in flood events by providing a foundation for real-time flood monitoring \cite{varga2021fast}. 

GNNs and their variants such as Graph Attention Networks (GAT) \cite{velickovic2017graph} are very successful in various fields but their capabilities in semantic segmentation, particularly for environmental monitoring such as flood detection, are still unexplored. Building upon existing advancements in GNNs, there exists a compelling research gap in the integration of GNNs with established architectures like U-Net for semantic segmentation. While U-Net and its variants have set benchmarks in medical and natural scene segmentation, the unique capabilities of GNNs to capture complex spatial relationships remain underutilized in these models \cite{relw2, relw3}. Consequently, our study integrates GNN with U-Net to take advantage of both and demonstrates the abilities of such architecture for flood segmentation.

\section{Background and Preliminaries} \label{sec:sota}

This section first describes U-Net as our approach is based on the U-Net architecture. Next, two state-of-the-art techniques, Efficient Neural Networks and SegFormer, which we compare in our analysis, are introduced. Moreover, our study examines the effect of transfer learning and model reprogramming with the SegFormer-based architecture.

\subsection{U-Net} \label{sec:unet}

U-Net is a CNN-based image segmentation model, initially developed by Olaf Ronneberger \cite{U-Net}. This model features an encoder-decoder architecture with a contraction path to capture image context and an expansion path for generating segmentation masks, forming a U-shaped structure. It is designed for efficient training with limited data and is suitable for deployment on edge devices due to its relatively lightweight design. In the encoder, input images undergo a series of convolutions, ReLU activations, and max pooling, creating deep feature maps that the decoder then upsamples, allowing feature reusability and gradient stability. The decoder culminates in 1x1 convolutions that classify pixels into segmentation classes. \cite{attention-u-net}.

\subsection{Efficient Neural Network} \label{sec:enet}

Efficient Neural Network (ENet), like U-Net, is also a deep learning architecture that was designed to provide a computationally efficient and accurate solution for real-time semantic segmentation tasks \cite{paszke2016enet}. Mehta et al. \cite{mehta2018espnet} proposed a modified ENet architecture. Their improved ENet architecture consists of a sequence of nineteen bottleneck encoders and four decoder blocks that are interconnected by skip connections. The encoder blocks gradually reduce the spatial resolution of the input image and increase the number of feature maps, while the decoder blocks upsample the feature maps and recover the spatial resolution of the image. The skip connections enable the network to fuse information from different levels of abstraction and improve the segmentation accuracy. The final layer is a 1x1 convolutional layer with a sigmoid activation function, which outputs a probability map for each pixel of the input image. 

\begin{figure*}[t] 
\centering

\includegraphics[width=0.99\textwidth]{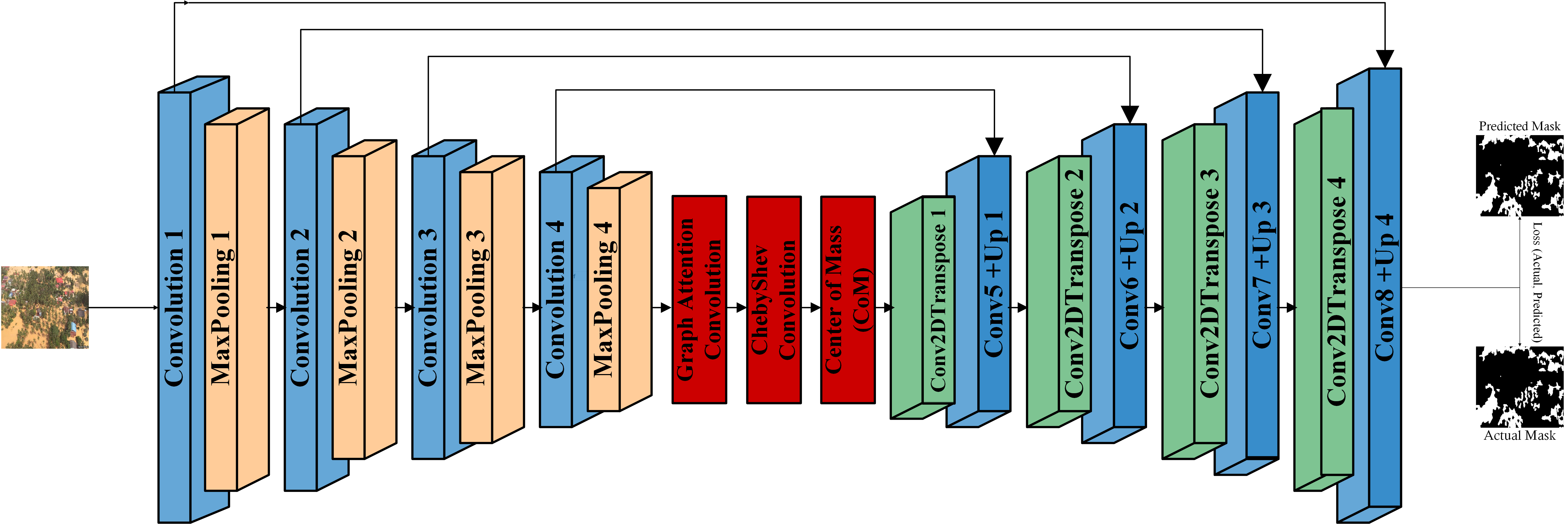}
\caption{\label{fig:arch}The proposed Graph Attention Convolutional U-NET (GAC-UNET) architecture consists of a series of convolutional layers followed by pooling layers in its encoder part. Then, the graph attention convolutional, Chebyshev graph convolutional, and center of mass layers are stacked between the encoder and decoder. The decoder also consists of a series of convolutional layers.}
\end{figure*}

\subsection{SegFormer} \label{sec:segformer}

SegFormer \cite{xie2021segformer} is a transformer-based neural network architecture proposed for image segmentation. Vision transformers introduced by Dosovitskiy et al. \cite{vit} have recently performed well in many computer vision tasks and they are the state-of-the-art in numerous computer vision tasks. SegFormer is inspired by vision transformers but it introduces characteristics for the segmentation tasks. It has an encoder decoder-style architecture: the encoder comprises hierarchical transformer blocks while the decoder consists of multi-layer perception (MLP) layers.

More specifically, the SegFormer encoder consists of hierarchically structured transformer blocks with self-attention which outputs multi-scale features. The encoder can be compared to that of U-Net which outputs multi-level features but instead of the convolutional layers, it has transformer blocks. The encoder, in addition to being hierarchical as mentioned above, removes positional encoding. Instead, SegFormer uses MixFFN blocks to preserve positional information. MixFFN block is a residual block consisting of two MLP layers and a single 3x3 convolution layer. The authors of SegFormer indicated that since semantic segmentation is a dense prediction task, positional encoding is not needed, and MixFFN blocks can propagate the positional information to the output.

As the decoder, SegFormer uses simple MLP layers. The intuition behind using MLP layers is that semantic segmentation has a dense representation in the output and hence it requires dense connections. The MLP layers are implemented as 1x1 convolution since the prediction is at the pixel level. The MLP decoder aggregates information from different scale outputs of the encoder, and hence it is capable of combining both local attention and global attention to render powerful representations. In addition to MLP layers, the decoder also has upsampling steps to match the scales of the multi-level representations.

\section{Methodology} \label{sec:method}

This section presents the proposed method, namely Graph Attention Convolutional U-NET (GAC-UNET), and the improvements made through transfer learning and model reprogramming.

\subsection{Graph Attention Convolutional UNET} \label{sec:improve}
We introduce an improved U-Net-like model that incorporates graph convolutional layers, specifically the Graph Attention Convolutional layer (GATConv) \cite{velickovic2018deep} and Chebyshev Convolutional layer (ChebConv) \cite{defferrard2016convolutional}, which are designed to enhance the model's ability to segment complex spatial patterns typically found in flooded areas. This architecture exploits the strengths of both traditional CNNs and advanced graph neural network techniques to improve semantic segmentation outcomes. The model, as shown in Figure \ref{fig:arch} consists of encoder-decoder U-NET-like architecture. It takes as input a colour image \( I \) of the flooded area with dimensions \( H \times W \times 3 \), where \( H \) and \( W \) are the height and width of the image, respectively, and 3 corresponds to the RGB channels. The corresponding grayscale ground truth \( H \times W \times 1 \) serves as the label. The encoder part of GAC-UNET consists of a series of convolutional and dilated convolutional layers.
These convolutional layers are followed by max-pooling layers that incrementally downsample the input image. This process ensures a broader receptive field and captures a richer representation of the input data which is essential for accurate segmentation. The convolution layer applies a kernel to the input using a convolution operation. Dilated convolutions introduce dilations in the kernel application to increase the receptive field without increasing the number of parameters which is essential for maintaining the size of the network and increasing its ability to learn from proceeding representations. 

Following the encoder, graph convolutional layers further process the feature maps. The first of these layers, the GATConv calculates attention coefficients between nodes by emphasizing features that are more important for the segmentation task. The introduction of GATConv layers at strategic points within the encoder allows for adaptive feature learning based on the topology of the data by utilizing attention mechanisms to weigh the importance of nodes (pixels) based on their contextual relevance in the graph structure. Wang et al. \cite{wang2019graph} have shown that GATConv is beneficial for handling the irregular and fragmented pattern seen in point cloud semantic segmentation, which is a somewhat similar task to flood image segmentation. The GATConv layer computes attention coefficients \( \alpha_{ij} \), which indicates the importance of features at node $j's$ to node $i$. These coefficients are computed using an attention mechanism that compares the features of the two nodes.
\begin{equation}
\alpha_{ij} = \frac{\exp\left(\text{LeakyReLU}\left(\mathbf{a}^T [\mathbf{W}h_i \| \mathbf{W}h_j]\right)\right)}{\sum_{k \in \mathcal{N}_i} \exp\left(\text{LeakyReLU}\left(\mathbf{a}^T [\mathbf{W}h_i \| \mathbf{W}h_k]\right)\right)}
\end{equation}
Here, \( \mathbf{a}^T \) is the learnable parameter vector of the attention mechanism, \( \mathbf{W} \) is a weight matrix applied to every node, \( h_i \) and \( h_j \) are the features of node \( i \) and \(j\), and \( \| \) denotes concatenation.

This GATConv layer is followed by the ChebConv layer, which employs Chebyshev polynomials to approximate the graph Laplacian, thus efficiently capturing higher-order interactions between nodes. This was motivated by Sahbi et al. \cite{sahbi2021learning} work which described that Chebyshev graph-based layers are very important for modelling complex spatial relationships that are not easily captured by traditional convolutional approaches. Specifically, Chebyshev convolution employs Chebyshev polynomials \( T_k(x) \) to compute the graph Laplacian's spectral approximation:
\begin{equation}
y_i = \sum_{k=0}^{K} T_k(\tilde{L}) x_i \theta_k
\end{equation}
where \( y_i \) is the output for node i, \( \tilde{L} \) is the scaled Laplacian, \( x_i \) is the input feature of node i, \( \theta_k \) are the parameters of the filter, and \( K \) is the order of the polynomial.

The ChebyConv layer is followed by a Center of Mass (CoM) layer as an intermediate layer that leverages the spatial distribution of features to enhance the localization of flooded areas by improving the model's precision and reliability. Hering et al. \cite{hering2022learn2reg} demonstrated that the Center of Mass (CoM) layer outperforms the Spatial Transformer Network (STN) in their applications. Motivated by their findings, we included the CoM layer in our network architecture, which computes centroids of deep features. This layer produces the final encoded features.

Next, the decoder reconstructs the segmentation map from the encoded features. It progressively restores the shape of feature maps while incorporating features from corresponding layers in the encoder via skip connections. Finally, a sigmoid activation function is applied to generate the predicted segmentation map.  The proposed architecture is shown in Figure \ref{fig:arch}.

\subsection{Transfer learning and model reprogramming} \label{sec:reprogram}

Transfer learning is widely used in various computer vision and other domains when limited data is available for network training. Commonly, transfer learning involves reusing a model developed for one task or a domain as the starting point for developing a model for a different task or a domain. Mostly, transfer learning involves retaining initial layers from the network (often a deep learning model) trained on the source domain and retraining or fine-tuning the final few layers of the network on the target domain. The intuition behind using transfer learning is that the initial network layers extract the low-level features of images that are common across the domains regardless of the image classes. Hence we can reuse these feature extractor parts of a model trained on a larger dataset. On the other hand, a few final layers need to be refined to better fit the task at hand.

Similarly, in flood segmentation, We too have a task where there is only a limited amount of data is available. Hence, here we evaluate whether transfer learning can help improve flood segmentation performance. For this task, we utilized the pre-trained model provided by the authors of SegFormer  \cite{xie2021segformer}. The model used in this particular case was trained on the Imagenet-1k dataset with a classification head intended for image classification \footnote{https://huggingface.co/nvidia/mit-b0}. The encoder part was kept the same while the decoder part was modified to suit semantic segmentation and fine-tuned by further training on the flood segmentation dataset.

Model reprogramming \cite{chen2022model} on the other hand is a very recent advancement in machine learning. Specifically, model reprogramming enables re-purposing a pre-trained model trained on the source domain for the target domain without fine-tuning it. In model reprogramming, the source model is kept frozen and an input transformation layer and an output mapping layer are introduced for re-purposing for the target domain. The purpose of the input transformation layer is to map the new input of the target domain to the input domain of the original (source domain) model. The purpose of the output mapping layer is to map the old outputs (source domain) to new outputs (target domain). Model reprogramming keeps the pre-trained model frozen and trains only the newly introduced input transformation and output mapping layers in an end-to-end manner. While this has shown to be effective in multiple tasks such as image classification, speech recognition, and biochemical sequencing \cite{chen2022model}, it is not yet examined in the context of semantic segmentation. 
We adapt the model reprogramming for the flood semantic segmentation task and examine its possibility to improve segmentation.

As a starting point source model for model reprogramming, we used the SegFormer model trained on ADE20k dataset for semantic segmentation  \footnote{https://huggingface.co/nvidia/segformer-b0-finetuned-ade-512-512}. This source model has 150 output classes. To repurpose this model for flood segmentation, we kept this model frozen (did not update any weights) but we introduced input transformation and output mapping. 
The input transformation employs an element-wise linear transformation taking each pixel value, applying a linear transformation, and outputting a corresponding new pixel value: 
\begin{equation}
    \Tilde{X}_{(m\times n)} = W_{(m\times n)} \cdot * X_{(m\times n)} + B_{(m\times n)}
    \label{eq:ip-transform}
\end{equation}
Here, $X$ is an $m \times n$ dimensional input image and $\Tilde{X}$ is the transformed input to be fed to the frozen network. The value of weights $W$ and biases $B$ are learned through backpropagation. The operation $\cdot*$ indicates elementwise multiplication. 

The output mapping layer is usually implemented as a fully connected layer before the softmax activation \cite{chen2022model}. However, since flood segmentation is a dense, pixel-level prediction task, we used a 2D 1x1 convolutional layer. It acts as a fully connected layer at the pixel level, connecting 150 channels of each old (source) output to the channels of the new output pixel.

\subsection{Loss functions} \label{sec:loss}


Two different loss functions, binary cross-entropy loss and Dice loss, are examined for the binary flood semantic segmentation task to investigate their ability to improve segmentation performance. 

\subsubsection{Binary cross-entropy loss} This is a commonly used loss function in semantic segmentation tasks \cite{reinke2021common}, where the goal is to classify each pixel in an image as either belonging to a particular class (foreground) or not (background). Binary cross-entropy loss is defined as:
\begin{equation}
	L_{BCE} = - \frac{1}{N} \sum_{i=1}^{N} y_i \cdot log(\hat{y_i}) + (1-y_i) \cdot log(1-\hat{y_i})
	\label{eq:bceloss}
\end{equation}
\noindent where $y_i$ is the ground truth label (either 0 or 1),  $\hat{y_i}$ is the predicted label (a value between 0 and 1 representing the probability of belonging to the foreground class), and $N$ is the number of samples.

\subsubsection{Dice loss} This function penalizes the dissimilarity and encourages similarity between the predicted segmentation mask and the ground truth mask, resulting in an accurately predicted mask \cite{reinke2021common}. It is the negative of the dice score. Dice loss is defined as follows:
\begin{equation}
	L_{Dice} = 1 -2 * \frac{\sum_{i=1}^{N} (y_i * \hat{y_i}) + \varepsilon} {\sum_{i=1}^{N} y_i + \sum_{i=1}^{N} \hat{y_i} + \varepsilon}
	\label{eq:dice}
\end{equation}
\noindent where $y_i$ and $\hat{y_i}$ are the true and predicted segmentation masks respectively. The $\varepsilon$ is a very small value added to prevent division by zero error. The sum is taken over all pixels in the segmentation. This loss function is widely used for image segmentation tasks, especially when there is a class imbalance issue \cite{reinke2021common}.  

\section{Evaluation and Results} \label{sec:results}

This section first provides a description of the dataset, the pre-possessing process, and the metrics used in the evaluation. Next, the results and the analysis are presented.
\subsection{Dataset} \label{sec:dataset}

\begin{figure}[t] 
\centering
\includegraphics[width=0.4\textwidth]{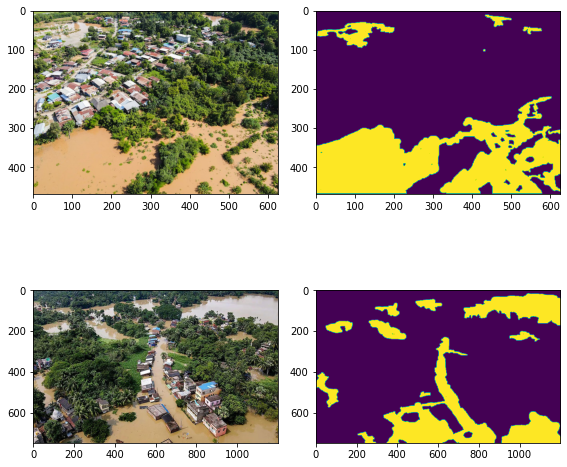}
\caption{\label{fig:sample}Sample images and corresponding masks from the dataset}
\end{figure}

The dataset used in the evaluation is Flood Area Segmentation dataset from kaggle \footnote{https://www.kaggle.com/datasets/faizalkarim/flood-area-segmentation}. This dataset consists of aerial photos of flooded areas. It has labelled masks indicating the water region. The dataset contains 290 images and annotated masks. The images are of varied dimensions. A few sample images with corresponding masks from the dataset are shown in Figure \ref{fig:sample}. The flooded areas are marked as 1 and the remaining pixels are marked as 0 in the label masks. Hence this dataset can be treated as a binary semantic segmentation dataset. There are 40.7\% positive class pixels and 59.3\% negative class pixels in the whole dataset.

The original dataset is split into training and testing subsets with a 70\%-30\% train-test split. The training dataset is expanded further with augmentations within the pre-processing pipeline (as described in the following subsection) to increase the number of training images and enhance the images with different properties to help the model generalize better.   


\subsection{Pre-processing}\label{sec:preprocess}

The flood-segmentation dataset consists of image-mask pairs in various sizes. However, all the considered segmentation networks expect images of a fixed size. Therefore, this RGB image size is set to 256 pixels in height and width dimensions. In our case, having a single channel output for the segmentation mask is sufficient as the fore-ground class ‘flooded areas’ is represented by 1s and the background class 'non-flooded areas' is represented as 0s. Moreover, since there is limited flood-segmentation data, the augmentation pipeline expands the dataset. 
Therefore, this study considered the following augmentation steps in a pre-processing pipeline to prepare the dataset for training the considered neural networks:

\begin{enumerate}
	\item Read RGB images as 3 channel float32 arrays and masks as single channel grayscale float32 arrays.
	\item Map all masks to 1s and 0s.
	\item Resize image-mask pairs to 512 x 512 in height and width dimensions using linear interpolation.
	\item Obtain five crop segments of size 256 x 256 from each image as four corner crops and one center crop.
	\item Repeat step 4 with the horizontally and vertically flipped image-mask pairs.
\end{enumerate}

Note that the augmentation is only used for the training set to create more samples and diversify the training data with the objective of increasing the models’ generalization performance. 

\begin{table*}[t!]
    \centering
    \caption{Comparison of proposed approach against baselines)}
    \label{tab:overall}
    \begin{tabular}{|p{0.5cm}|p{7.5cm}|p{2cm}|p{2cm}|p{2cm}|}
        \toprule
        \textbf{Sr. No} & \textbf{Model} & \textbf{Dice Score (Out of 1.00)} & \textbf{Intersection Over Union (Out of 1.00)} & \textbf{mean Average Precision (Out of 1.00)} \\
        \midrule
        1 & Convolutional U-NET with BCE & 0.82 & 0.71 & 0.75  \\ \midrule
        2 & Convolutional U-NET with DICE & 0.83 & 0.73 & 0.76  \\ \midrule
        3 & Efficient Neural Network with BCE & 0.79 & 0.69 & 0.78  \\ \midrule
        4 & Efficient Neural Network with DICE & 0.82 & 0.71 & 0.78  \\ \midrule
        5 & Vision Transformer based SegFormer & 0.79 & 0.67 & 0.72  \\ \midrule
        6 & Vision Transformer based SegFormer (Transfer Learning) & 0.86 & 0.76 & 0.82  \\ \midrule
        7 & Vision Transformer based SegFormer (Model Reprogramming) & 0.80 & 0.69 & 0.75  \\ \midrule
        8 & Graph Attention Convolutional U-NET with BCE (ours) & 0.83 & 0.79 & 0.77  \\ \midrule
        9 & Graph Attention Convolutional U-NET with DICE (ours)& \textbf{0.94} & \textbf{0.89} & \textbf{0.91}  \\ 
        \bottomrule
    \end{tabular}
    \label{tab:gat-results}
\end{table*}

\subsection{Metrics}


For the evaluation of the proposed approach, the three commonly used metrics from the segmentation domain are employed: Intersection over Union, Dice Score, and Mean Average Precision. The following paragraphs introduce the three metrics.

\subsubsection{Intersection over Union (IoU)} IoU, also known as the Jaccard index, measures the similarity between the predicted segmentation mask $Y_{pred}$ and the ground truth mask $Y_{true}$. The IoU or Jaccard index is calculated as the ratio of the intersection of the predicted mask and ground truth mask to the union of the predicted mask and ground truth mask.
\begin{equation}
	IOU = \frac{| Y_{true} \cap Y_{pred}|}{| Y_{true} \cup Y_{pred}|}
	\label{eq:iou}
\end{equation}
The IoU index ranges from 0 to 1, with 1 indicating a perfect overlap between the predicted mask and the ground truth mask. In other words, the higher the IoU index, the better the performance of the segmentation model.

\subsubsection{Mean Average Precision (mAP)} This metric measures the precision of a model at different Intersections over Union (IoU) thresholds and then averages these precision scores to compute the final mAP score.

\subsubsection{Dice Coefficient } This metrics measures the overlap between the predicted segmentation mask and the ground truth mask \cite{aziz2024global}. The Dice coefficient, also known as the Dice index, ranges from 0 to 1, with 1 indicating a perfect overlap between the predicted and ground truth masks. It is calculated as follows:
\begin{equation}
	Dice = \frac{2 * | Y_{true} \cap Y_{pred}|}{| Y_{true}| + |Y_{pred}|}
	\label{eq:dice2}
\end{equation}

\subsection{Results and Analysis} \label{sec:transfer-results}

\begin{figure}[t] 
\centering
\includegraphics[width=0.5\textwidth]{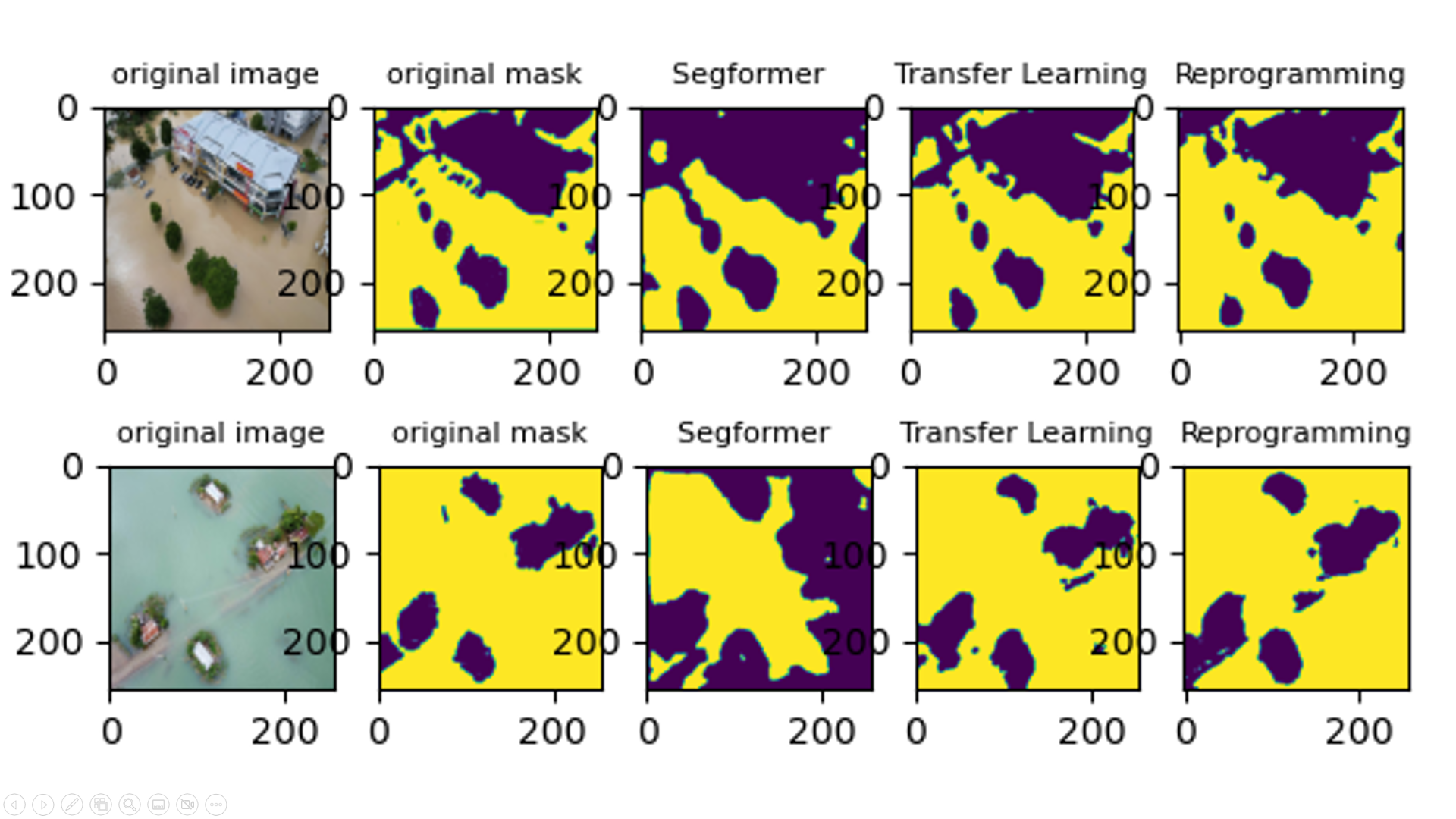}
\caption{\label{fig:timeline} Output comparison of traditionally trained SegFormer, SegFormer with transfer learning, and segformer with model reprogramming }
\end{figure}

Table \ref{tab:gat-results} compares the proposed Graph Attention Convolutional U-NET (GAC-UNET) with several state-of-the-art approaches as well as examines the impact of the transfer learning and model reprogramming. Convolutional U-NET, Efficient Neural Network, and GAC-UNET are evaluated under two settings: with Binary Cross-Entropy (BCE) and with Dice loss (DICE) to assess the impact of Dice loss compared to BCE. Whereas, SegFormer and its variants are evaluated to study the effect of transfer learning and model reprogramming. Convolutional U-NET with BCE loss achieved a 0.82 Dice score, 0.71 IoU, and 0.75 mAP. However, when trained with the Dice loss function, accuracy increased to 0.83 in terms of the Dice metric, 0.73 IoU, and 0.75 mAP. This indicates that Dice loss improves segmentation even for the traditional convolutional U-NET. The same comparison was carried out for Efficient Network. With BCE loss, it achieved a 0.79 Dice score, 0.69 IoU, and 0.78 mAP. However, when trained on the Dice loss, the metrics improved to a Dice score of 0.82, IoU of 0.71, and mAP of 0.78. As with convolutional U-NET, for Efficient Network, Dice loss function reduced the discrepancy between the predicted and ground truth maps.


The same Table \ref{tab:gat-results} also includes the examination of transfer learning and model reprogramming to the SegFormer model. This architecture was selected as the pre-trained SegFormer model from different domains is available. Additionally, two sample test images are shown in Figure \ref{fig:timeline}, for visual comparison. Transfer learning notably improved the performance of the SegFormer, as is evident from all three evaluation metrics. With transfer learning the SegFormer model achieved 0.86 Dice score, 0.76 IoU and 0.82 mAP outperforming all three traditionally trained state-of-the-art models -- convolutional U-NET, Efficient Neural Network, and Vision Transformer-based SegFormer -- as observed in Table\ref{tab:gat-results}. Model reprogramming on the other hand was only able to improve the performance marginally compared to traditional training. While transfer learning still improved the performance, the improvement was lower compared to that achieved by transfer learning. Hence, based on these empirical results, we can observe that while model reprogramming is applicable in the semantic segmentation context, it is not superior to transfer learning.

However, it is worth noting some additional benefits of model reprogramming. It has fewer number of trainable parameters and, hence is more resource efficient. Morel reprogramming keeps the original model unchanged, so it can still be used for the original task. Moreover, multiple similar tasks (different semantic segmentation tasks) can be solved by one base (source) model and multiple input layers and output mapping layers in a plug-and-play manner. This is more memory efficient than having a separate model for each task.


Table \ref{tab:gat-results} also shows the result for the proposed GAC-UNET including two variants: with BCE and Dice loss. Yet again, the same as with convolutional U-NET and Efficient Neural Network, Dice loss improves the performance in comparison to using BCE loss in terms of all tree performance metrics. Comparing convolutional U-Net with our GAC-UNET, it can be observed that for both loss functions, the introduction of new layers -- graph attention convolution, Chebyshev, and center of mass -- improved the performance of the U-Net architecture.

\begin{figure}[t] 
\centering
\includegraphics[width=0.5\textwidth]{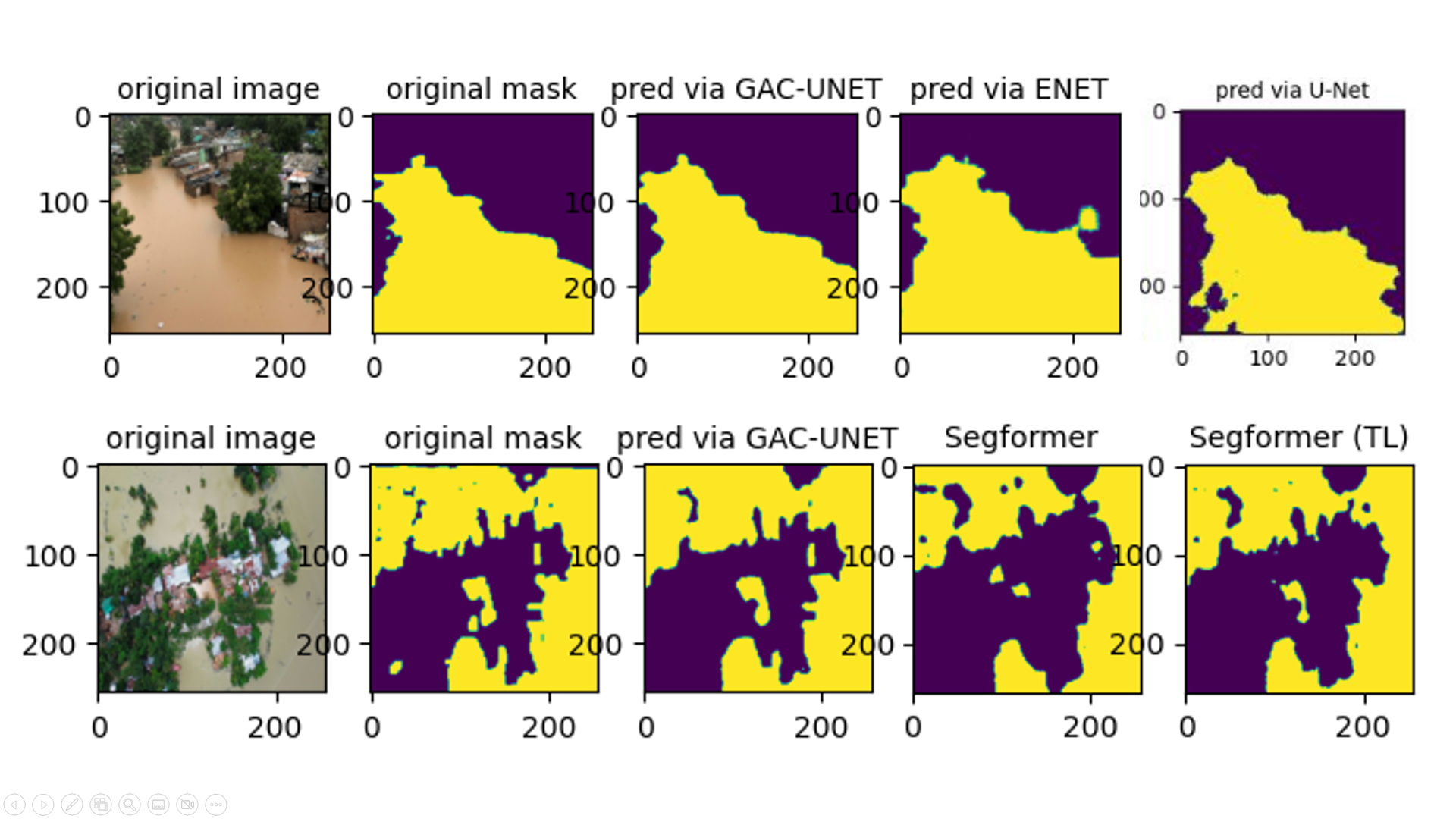}
\caption{\label{fig:compare} Output comparison of our proposed model with baselines (U-Net, E-net,  SegFormer, and SegFormer with transfer learning)}
\end{figure}


For visual comparison, Figure \ref{fig:compare} shows two examples of flood images with corresponding segmentation masks obtained using different techniques. It is clear that GAC-UNET is able to capture the flooded area map very closely to the ground truth.

Overall, as observed from Table \ref{tab:gat-results}), the proposed GAC-UNET outperforms other considered techniques in terms of DICE score (0.94), IoU score (0.89), and mAP score (0.91) on flood segmentation task. This indicates that the proposed model provides more accurate and precise segmentation results compared to the other evaluated models.

\section{Conclusion} \label{sec:conclusion}

Determining the extent of the flood is vital for decision-making and infrastructure planning in flood-prone regions and can be done through semantic segmentation. This paper introduces Graph Attention Convolutional U-NET, a semantic segmentation model based on graph attention mechanisms and Chebyshev convolutional layers, designed for the identification of flooded areas. The proposed model was evaluated against existing segmentation architectures and showed superior performance. It achieved a mean Average Precision of 0.91, a Dice score of 0.94, and an Intersection over Union of 0.89. These results not only indicate an improvement over existing segmentation methods but also highlight the effectiveness of integrating graph-based neural network techniques for processing complex spatial data.

Furthermore, this paper also investigates the possibility of utilizing ML techniques such as transfer learning and model reprogramming for the identification of flooded areas. Empirical results using SegFormer as a baseline affirmed the benefits of transfer learning and model reprogramming in enhancing model performance under constrained data conditions, which is typical in flooded area detection scenarios. Although model reprogramming showed better results than the base model, it did not outperform transfer learning. 

Future work will investigate refining and combining these approaches for broader applicability in real-time scenarios and extending the model to encompass other forms of environmental segmentation tasks.

\bibliographystyle{IEEEtranN}
\bibliography{references}

\end{document}